\documentclass[journal]{IEEEtran}
\usepackage{amsmath,amsfonts}
\usepackage{array}
\usepackage{hyperref}
\usepackage{stfloats}
\usepackage{url}
\usepackage{verbatim}
\usepackage{graphicx}
\usepackage{cite}
\usepackage{graphicx}
\usepackage{url}
\usepackage{cleveref}
\usepackage{booktabs}
\usepackage{algorithm}
\usepackage{algpseudocode}
\usepackage{multirow}
\usepackage{bm}
\usepackage{colortbl}
\usepackage[table]{xcolor}
\usepackage{subfigure}
\usepackage{lipsum} 
\usepackage{amssymb}

\usepackage{xspace}
\usepackage{pifont}
\usepackage{siunitx}
\newcommand{\eg}{\textit{e.g.}\@\xspace}

\usepackage{enumitem}
\usepackage{adjustbox}

\begin{document}

\title{BitC-3DGS: High-Capacity 3DGS Watermarking via Bit Compression}

\author{
    Yuquan~Bi, Baosheng~Yu, Yingke~Lei, Jianwei~Yang, Hongsong~Wang, Jie~Gui,~\IEEEmembership{Senior Member,~IEEE,} Yuan~Yan~Tang,~\IEEEmembership{Life Fellow,~IEEE,} and James~Tin-Yau~Kwok,~\IEEEmembership{Fellow,~IEEE}%

\thanks{\textit{Corresponding author: Jie Gui.}}
\thanks{Y. Bi is with the School of Cyber Science and Engineering, Southeast University, Nanjing, China (e-mail: yuquanbi@seu.edu.cn).}%
\thanks{B. Yu is with the Lee Kong Chian School of Medicine, Nanyang Technological University, Singapore (e-mail: baosheng.yu@ntu.edu.sg).}%
\thanks{Y. Lei is with National University of Defense Technology, Hefei 230037, China (e-mail: leiyingke@163.com).}%
\thanks{J. Yang is with the Institute of AI for Industries, Chinese Academy of Sciences, Nanjing, China (e-mail: yangjianwei20@mails.ucas.ac.cn).}%
\thanks{H. Wang is with School of Computer Science and Engineering, Key Laboratory of New Generation Artificial Intelligence Technology and Its Interdisciplinary Applications, Ministry of Education, Southeast University, Nanjing 210096, China (e-mail: hongsongwang@seu.edu.cn).}%
\thanks{J. Gui is with the School of Cyber Science and Engineering, Southeast University, and also with Purple Mountain Laboratories, and also with Engineering Research Center of Blockchain Application, Supervision And Management (Southeast University), Ministry of Education, Nanjing 210000, China (e-mail: guijie@seu.edu.cn).}%
\thanks{Yuan Yan Tang is with the Department of Computer and Information
Science, University of Macau, Macau, China, and also with Faculty of Science and Technology, UOW College Hong Kong, Hong Kong, China (e-mail: yytang@um.edu.mo).}%
\thanks{J. T. -Y. Kwok is with the Department of Computer Science and Engineering, The Hong Kong University of Science and Technology, Hong Kong, China (e-mail: jamesk@cse.ust.hk).}%
}



\maketitle

\begin{abstract}
High-capacity watermarking is necessary for 3D Gaussian Splatting (3DGS) assets to embed rich information (\eg, ownership, provenance, and authentication codes), enabling reliable identification and provenance tracking in large-scale 3D asset pipelines. Existing bit-to-token watermarking methods based on a pre-trained text encoder are limited to 77-bit messages due to CLIP's fixed 77-token context length, as tokens beyond this limit are unsupported by learned positional embeddings. To address this limitation, we introduce BitC-3DGS, a bit-compression framework that encodes multiple message bits per token. It employs a bit-compressed tokenization scheme that encodes multiple bits within the same chunk into a single semantic token. To enable recovery of the compressed information, it further introduces a dual-branch architecture for joint chunk decompression and bit decoding, along with a hard-message sampling strategy to improve combinatorial coverage during decoder training. Extensive experiments on the Blender and LLFF datasets demonstrate the effectiveness of BitC-3DGS for high-capacity watermarking, achieving high message recovery accuracy and rendering fidelity. For example, it supports 128-bit message capacity with recovery accuracy comparable to that of 64-bit messages in recent state-of-the-art methods.
\end{abstract}

\begin{IEEEkeywords}
3D Gaussian splatting, digital watermarking, copyright protection.
\end{IEEEkeywords}

\section{Introduction}
\label{sec:intro}

\IEEEPARstart{N}{eural} fields, particularly 3D Gaussian Splatting (3DGS), provide an efficient representation for photorealistic, real-time novel view synthesis~\cite{nerf,3dgs}. They enable large-scale creation, sharing, and deployment of 3D assets for applications such as augmented and virtual reality, digital twins, and robotics. This rapid adoption introduces challenges in ownership protection, provenance tracking, and responsible reuse of 3D content~\cite{luo2026mantlemark,huang2023robust}. High-capacity watermarking has thus become an important capability for 3DGS~\cite{gaussianmarker,3d-gsw,guardsplat}. It enables rich and verifiable metadata, including ownership credentials, licensing information, and content lineage, to be embedded directly within the representation while preserving rendering quality and efficiency. By distributing structured information across many Gaussians and their attributes, such methods improve robustness to common processing operations, including pruning, densification, relighting, compression, quantization, and repeated rendering and re-capture. These properties support scalable asset governance and enhance traceability across realistic 3D content production and distribution pipelines.

\begin{figure}[t]
\centering
\includegraphics[width=1\columnwidth]{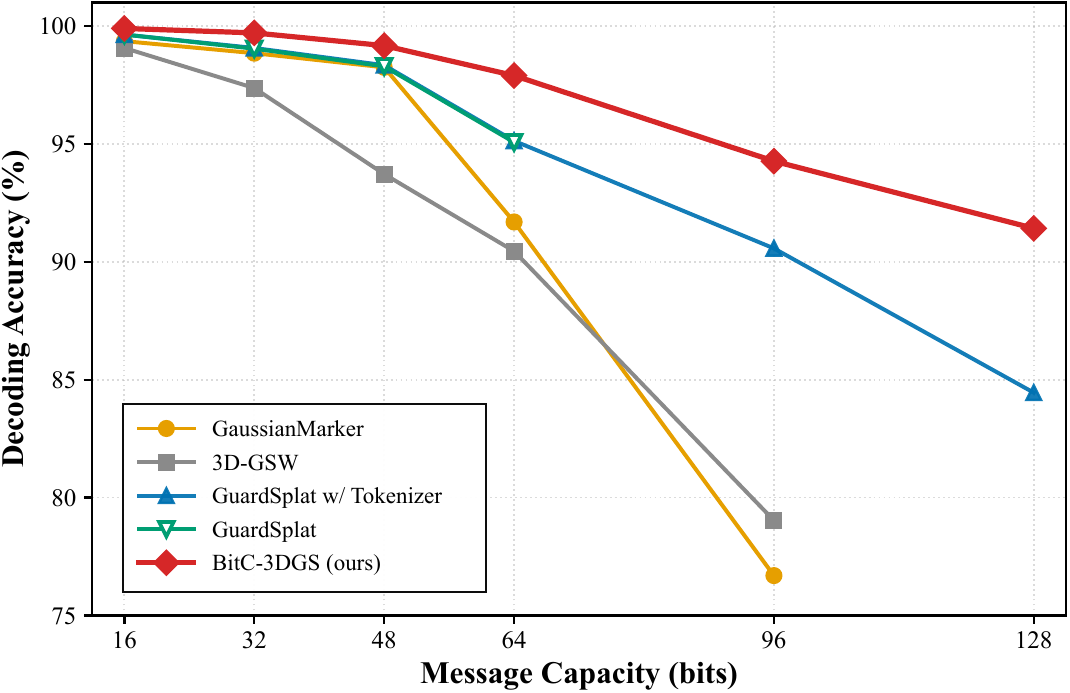} 
\par\vspace{-2mm}
\caption{\textbf{Decoding accuracy versus message capacity.} Conventional image-space watermarking methods operating on low-level signals (\eg, GaussianMarker~\cite{gaussianmarker} and 3D-GSW~\cite{3d-gsw}) suffer notable performance degradation as message capacity increases. Semantic watermarking based on bit-to-token encoding (\eg, GuardSplat~\cite{guardsplat}) improves decoding accuracy but is inherently limited to 77-bit messages. In contrast, the proposed BitC-3DGS enables watermark messages exceeding 77 bits while preserving robust decoding accuracy. For example, it supports a 128-bit message capacity with decoding bit accuracy comparable to that of 64-bit settings in prior methods such as GaussianMarker and 3D-GSW.}
\label{fig_intro_capcity}
\end{figure}

As shown in Fig.~\ref{fig_intro_capcity}, conventional image-space watermarking methods such as GaussianMarker~\cite{gaussianmarker} and 3D-GSW~\cite{3d-gsw} degrade rapidly as the message length increases, since limited perturbation budgets must encode increasingly dense low-level watermark signals, making reliable recovery difficult. Semantic watermarking methods, represented by GuardSplat~\cite{guardsplat}, address this limitation by shifting watermark embedding from low-level perturbations to high-level semantic alignment. However, they rely on pre-trained text encoders, and CLIP supports only a fixed 77-token context via learned positional embeddings~\cite{clip}. As a result, the current bit-to-token scheme does not scale beyond 77-bit messages, posing the challenge of encoding substantially more information within a fixed number of semantic tokens.

To address this limitation, we introduce BitC-3DGS, a bit-compression framework for high-capacity watermarking. BitC-3DGS employs bit-compressed tokenization to group multiple bits into a single semantic token: a binary message is split into fixed-length chunks of $n$ bits, each chunk represents one of $2^n$ possible values, and is deterministically mapped to tokens via a position-aware lookup table built from the CLIP vocabulary. This replaces bit-level encoding with a chunk-level semantic representation, enabling higher watermark capacity under the same encoder budget. However, as shown in Fig.~\ref{fig_intro_capcity}, this compression increases decoding difficulty, since bit-level recovery must account for the chunk structure. For example, while bit-compressed tokenization enables GuardSplat~\cite{guardsplat} to embed more bits, it still results in moderate decoding accuracy. To address this issue, we design a dual-branch decoder that jointly performs chunk-level and bit-level prediction. The chunk branch predicts the discrete chunk index among $2^n$ candidates, while the bit branch directly predicts each individual bit, as in standard bit-to-token methods. This joint optimization enhances robust reconstruction of compressed watermark messages.

While the payload space in semantic watermarking grows exponentially with payload size, decoder training is often limited by insufficient combinatorial coverage. In practice, conventional random sampling explores only a small subset of the message space, leading to poor coverage in high-capacity settings. When this sampled subset remains fixed throughout training, the decoder repeatedly encounters the same message patterns, which can lead to overfitting and weaken recovery of newly assigned messages. To mitigate this, we replace naive random sampling with a hard-message sampling strategy that prioritizes informative and diverse message patterns, so each training example spans a broader region of the combinatorial message space. Specifically, we maintain a message buffer and an online accuracy memory to guide sampling. After each iteration, incorrectly decoded messages are recorded, and their historical accuracies are updated via exponential smoothing. At each epoch, the buffer is rebuilt by selecting low-accuracy messages as hard samples and supplementing them with unseen messages to preserve diversity. The proportion of hard samples gradually increases until the training set consists entirely of the most difficult messages, enabling progressive coverage of hard cases and improving generalization to unseen messages.

Our main contributions are as follows:
\begin{itemize}
\item We propose BitC-3DGS, a bit-compression framework for high-capacity watermarking that mitigates the 77-token limitation of CLIP-based semantic encoding. 
\item We introduce a dual-branch decoding architecture, together with a hard-message sampling strategy, to enable reliable training under high-capacity settings. 
\item We conduct extensive experiments on two widely used datasets, Blender and LLFF, and the results demonstrate the effectiveness of the proposed framework for high-capacity 3DGS watermarking.
\end{itemize}

\section{Related Work}
\label{sec:related_work}

\subsection{2D Digital Watermarking}
2D digital watermarking embeds hidden information into images or videos in an imperceptible yet recoverable manner for copyright protection. Early methods rely on spatial- or frequency-domain perturbations \cite{tradition-wm1,tradition-wm2,tradition-wm3,tradition-wm4,tradition-wm5}, but are often vulnerable to common distortions such as compression and cropping \cite{related4,tradition-wm1}. Recent approaches are predominantly learning-based, where neural networks jointly learn embedding and extraction functions, with representative methods including HiDDeN \cite{hidden}, CIN \cite{cin}, and StegaStamp \cite{stegastamp}, achieving improved robustness and imperceptibility \cite{luo2025dvmark,guo2026sacmark}. Extensions further consider challenging settings such as encrypted and compressed domains, watermark removal attacks, and screen-shooting robustness \cite{haddad2020joint,fang2019screenshooting,chen2025imprints,guo2025robust}. More recently, diffusion-based watermarking has been explored to protect generated content from models such as DDPM \cite{diff-watermark1,diff-watermark2,diff-watermark3,ddpm,related8}. However, these methods remain designed for 2D signals and do not account for the structural representation, rendering process, or editing flexibility of modern 3D assets.

\subsection{NeRF Watermarking}
With advances in neural implicit representations~\cite{lu2025hacc}, including neural radiance fields (NeRF)~\cite{nerf} and their variants~\cite{related1,related2,related3}, watermarking has been extended to implicit 3D representations. Representative methods such as CopyRNeRF \cite{copyrnerf} and StegaNeRF \cite{steganerf} embed messages into radiance field parameters via optimization, while WateRF \cite{waterf} introduces frequency-domain perturbations to enhance robustness across radiance-field variants. These approaches demonstrate the feasibility of embedding watermarks in neural 3D scene representations beyond 2D content. However, NeRF-based watermarking is constrained by the implicit nature of radiance fields, which makes control over watermarks, spatial consistency, and editing less tractable. These limitations motivate the exploration of more explicit and editable 3D representations, such as 3D Gaussian Splatting, for practical watermarking.

\subsection{3DGS Watermarking}
3D Gaussian Splatting (3DGS)\cite{3dgs} has emerged as a widely used representation for editable 3D content, enabling applications in virtual reality\cite{3dgs_human1,3dgs_human2,3dgs_human3,3dgs_human4}, scene reconstruction~\cite{3dgs_recon1,3dgs_recon2,3dgs_recon3,3dgs_recon4}, and generative pipelines~\cite{3dgs_gen1,3dgs_gen2,3dgs_gen3,3dgs_gen4}. Its growing adoption has motivated the development of watermarking methods tailored to 3DGS and its variants~\cite{gaussianmarker,gs-hider,3d-gsw,guardsplat,instantsplamp,li2026gaussianseal}. Existing approaches differ in how watermark information is embedded into and recovered from the representation. GaussianMarker~\cite{gaussianmarker} constrains embedding via uncertainty-aware parameter modulation, while 3D-GSW~\cite{3d-gsw} uses frequency-guided optimization on rendered views. These methods rely on low-level signal modifications and tend to degrade as payload size increases. GuardSplat~\cite{guardsplat} instead introduces a semantic formulation with CLIP-guided supervision and SH-coefficient perturbations, improving robustness over pixel-level designs. However, its one-bit-per-token strategy is limited by the fixed CLIP context length, imposing a hard capacity ceiling, and training on a fixed message subset further biases decoding toward seen patterns as the message space grows. These limitations motivate high-capacity semantic watermarking for 3DGS beyond fixed token budgets and static training subsets.

\begin{figure*}[t]
\centering
\includegraphics[width=\linewidth]{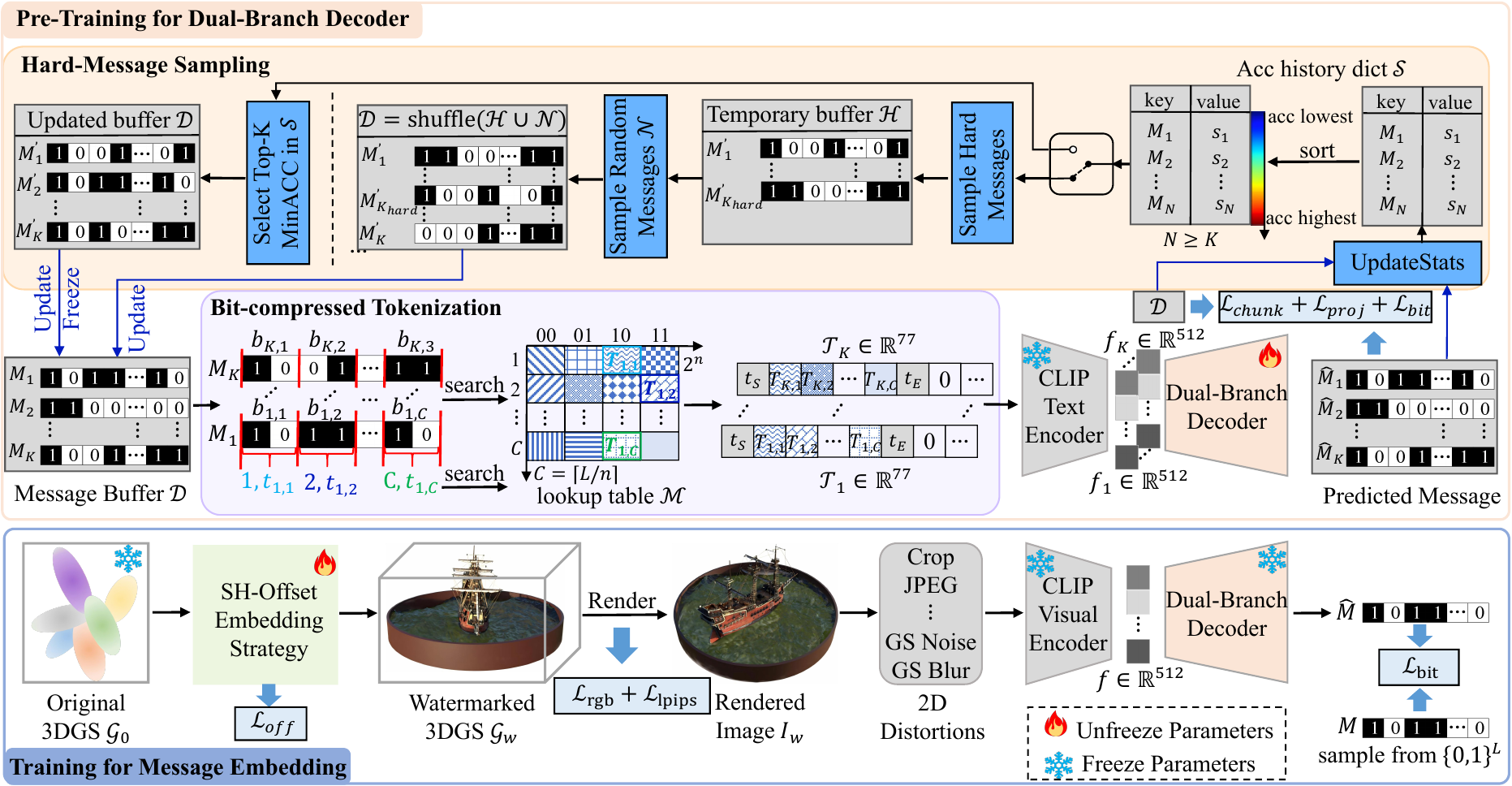}
\par\vspace{-2mm}
\caption{\label{fig_architecture} \textbf{Overview of the proposed  BitC-3DGS framework.} The method contains two stages. Stage I (decoder pre-training): Messages sampled from the buffer $\mathcal{D}$ are converted into compact token sequences via bit-compressed tokenization, encoded by the pre-trained CLIP text encoder, and used to train the dual-branch decoder $\mathcal{D}_M$. During this stage, a hard-message sampling strategy continually updates $\mathcal{D}$ by retaining historically difficult messages and introducing unseen ones to mitigate fixed-subset bias. Stage II (watermark embedding): The pretrained decoder $\mathcal{D}_M$ is frozen and used to supervise watermark embedding into the 3DGS representation. Rendered views with optional distortions are encoded by the CLIP image encoder and decoded by $\mathcal{D}_M$, while 3DGS parameters are optimized using image reconstruction, bit recovery, and embedding regularization losses.}
\end{figure*}

\section{Method}
\label{sec:method}

This section presents the BitC-3DGS framework. We first describe the 3DGS watermarking pipeline, including decoder pretraining and watermark embedding, and then introduce the key components: bit-compressed tokenization, dual-branch decoding, and hard-message sampling.

\subsection{Overview}

Fig.~\ref{fig_architecture} illustrates the proposed BitC-3DGS framework, which consists of decoder pre-training and watermark embedding. In the first stage, binary messages are tokenized and encoded by a frozen CLIP text encoder to train the message decoder. In the second stage, the message is embedded into a 3DGS asset by optimizing SH-coefficient offsets under the supervision of the frozen decoder. The embedded message is subsequently recovered from rendered images using the CLIP image encoder and the decoder's bit branch.

In the decoder pre-training stage, binary messages sampled from the buffer $\mathcal{D}$ are converted into compact token sequences via bit-compressed tokenization and encoded by the pre-trained CLIP text encoder $\mathcal{E}_T$ into 512-dimensional embeddings. These embeddings are then fed into the dual-branch decoder $\mathcal{D}_M$, which is trained with a combination of the chunk-level loss $\mathcal{L}_{\mathrm{chunk}}$, projected-bit loss $\mathcal{L}_{\mathrm{proj}}$, and direct-bit loss $\mathcal{L}_{\mathrm{bit}}$:
\begin{equation}
\mathcal{L}_{\mathrm{decoder}}
=
\lambda_s \mathcal{L}_{\mathrm{chunk}}
+
\lambda_p \mathcal{L}_{\mathrm{proj}}
+
\lambda_b \mathcal{L}_{\mathrm{bit}},
\end{equation}
where $\lambda_s$, $\lambda_p$, and $\lambda_b$ weight the three losses defined in Section~\ref{sec:decoder}. The buffer $\mathcal{D}$ is dynamically updated each epoch via a hard-message sampling strategy that aggregates challenging samples from previous training iterations.

In the watermark embedding stage, the pre-trained decoder $\mathcal{D}_M$ supervises the optimization of watermark injection into the 3DGS representation via multi-view rendering. Given a rendered image $I_w$, random visual perturbations are applied, and the result is encoded by the pre-trained CLIP image encoder $\mathcal{E}_I$ before being passed to the decoder to obtain the predicted message $\hat{M} = \mathcal{D}_M\!\left(\mathcal{E}_I(I_w)\right)$.
The SH-coefficient offsets are optimized using the direct-bit loss $\mathcal{L}_{\mathrm{bit}}$, image fidelity losses, and offset regularization $\mathcal{L}_{\mathrm{off}}$~\cite{guardsplat}:
\begin{equation}
\mathcal{L}_{\mathrm{watermark}}
=
\lambda_{\mathrm{bit}}\mathcal{L}_{\mathrm{bit}}
+
\lambda_{\mathrm{image}}(\mathcal{L}_{\mathrm{rgb}} + \mathcal{L}_{\mathrm{lpips}})
+
\lambda_{\mathrm{off}}\mathcal{L}_{\mathrm{off}},
\end{equation}
where $\lambda_{\mathrm{bit}}$, $\lambda_{\mathrm{image}}$, and $\lambda_{\mathrm{off}}$ control the contribution of each term. The image fidelity term consists of an RGB reconstruction loss and a perceptual LPIPS loss~\cite{lpips}. Specifically, the RGB reconstruction loss is defined as
\begin{equation}
\mathcal{L}_{\mathrm{rgb}}
=
\lambda_{\mathrm{ssim}}\mathcal{L}_{\mathrm{ssim}}(I_w,I_o)
+
(1-\lambda_{\mathrm{ssim}})\mathcal{L}_{1}(I_w,I_o),
\end{equation}
where $I_o$ denotes the image rendered from the original 3DGS model at the same camera viewpoint as $I_w$.

\subsection{Bit-Compressed Tokenization}
\label{sec:tokenizer}

To address the fixed token limit, we introduce bit-compressed tokenization, which encodes multiple bits into a single token. Given a binary message $M_k$ of length $L$, it is first split into $C=\lceil L/n\rceil$ non-overlapping $n$-bit chunks:
\begin{equation}
M_k = [b_{k,1}, b_{k,2}, \ldots, b_{k,C}],
\end{equation}
each mapped to an index $t_{k,i} \in \{0,\ldots,2^n-1\}$:
\begin{equation}
t_{k,i} = \sum_{j=1}^{n} b_{k,i,j} \cdot 2^{n - j}.
\end{equation}
A position-aware lookup table is constructed from the CLIP vocabulary after excluding reserved token IDs. The resulting vocabulary $\mathcal{V}_{\mathrm{valid}}$ is randomly permuted as $[p_1,\ldots,p_N]$. The lookup table  $\mathcal{M} \in \mathbb{N}^{C \times 2^n}$ is defined as
\begin{equation}
\mathcal{M}[i, j] = p_{(i-1)\cdot 2^n + j}, \quad \text{for } i \in [1,C],\ j \in [1,2^n],
\end{equation}
Following \cite{guardsplat}, the final token sequence is formed by concatenating the start token $t_S$, mapped tokens, and end token $t_E$:
\begin{equation}
\mathcal{T}_k=
[t_S,\; \mathcal{M}[1,t_{k,1}+1],\; \ldots,\; \mathcal{M}[C,t_{k,C}+1],\; t_E].
\end{equation}
The sequence is zero-padded to 77 tokens for CLIP compatibility. Since $t_S$ and $t_E$ occupy two of the 77 positions, the one-bit-per-token scheme supports at most 75 bits, whereas bit-compressed tokenization raises this bound to $75n$ bits. This chunk-to-token design enables higher-capacity watermarking under a fixed token budget. For simplicity, this bit-compressed tokenization process is denoted as $\texttt{Tokenizer}(M_k,\mathcal{M})$.

\begin{algorithm}[t]
\renewcommand{\algorithmicrequire}{\textbf{Input:}}
\renewcommand{\algorithmicensure}{\textbf{Output:}}
\caption{Decoder Pre-training}
\label{alg:train_overview}
\begin{algorithmic}[1]
\State \textbf{Input:} Message length $L$, compression rate $n$, total epochs $E$, freeze epoch $E_{\text{freeze}}$, maximum pool size $K$, threshold $\sigma$, initial hard ratio $\tau_0$, warm-up slope $\alpha$
\State \textbf{Require:} Message buffer $\mathcal{D}$, Accuracy history dict $\mathcal{S}$, pretrained CLIP text encoder $\mathcal{E}_T$, static lookup table $\mathcal{M}$
\State \textbf{Output:} Trained decoder $\mathcal{D}_M$

\State \textbf{Initialize} $\mathcal{S} \leftarrow \emptyset$, $\mathcal{M}$, and $\mathcal{D}$ with $K$ random binary messages of $L$

\For{epoch $e = 1$ to $E$}
    \State $\mathcal{T} \leftarrow \emptyset$
    \For{$M_k \in \mathcal{D}$}
        \State $\mathcal{T}_k \gets \texttt{Tokenizer}(M_k, \mathcal{M})$ \Comment{Sec. \ref{sec:tokenizer}}
        \State $\mathcal{T} \gets \mathcal{T} \cup \{(\mathcal{T}_k, M_k)\}$
    \EndFor
    
    \For{each batch $(\mathcal{T}_k, M_k)$ sampled from $\mathcal{T}$}
        \State $f \gets \mathcal{E}_T(\mathcal{T}_k)$ 
        \State $\hat{M}_k, \tilde{M}_k, s \gets \mathcal{D}_M(f)$ \Comment{Sec. \ref{sec:decoder}}
        \State $\mathcal{L} \gets \lambda_s\mathcal{L}_{\mathrm{chunk}} +\lambda_p\mathcal{L}_{\mathrm{proj}} +\lambda_b\mathcal{L}_{\mathrm{bit}}$
        \State Backpropagate and update $\mathcal{D}_M$
        \State $\mathcal{S} \gets \texttt{UpdateStats}(\mathcal{S}, M_k, \hat{M}_k)$ \Comment{Sec. \ref{sec:hgdr}}
    \EndFor
    
    \If{$e < E_{\text{freeze}}$}
        \State $r_{\text{hard}} \gets \min(1.0, \tau_0+\alpha \cdot e)$
        \State $K_{\text{hard}} \gets \lfloor r_{\text{hard}} \cdot K \rfloor$
        \State $\mathcal{H} \gets \texttt{SampleHardMessages}(\mathcal{S}, \sigma, K_{\text{hard}})$
        \State $\mathcal{N} \gets \texttt{SampleRandomMessages}(K - |\mathcal{H}|)$
        \State $\mathcal{D} \gets \texttt{Shuffle}(\mathcal{H} \cup \mathcal{N})$
    \ElsIf{$e = E_{\text{freeze}}$}
        \State $\mathcal{D} \gets \texttt{Shuffle}(\texttt{TopKMinAccuracy}(\mathcal{S}, K))$
    \Else
        \State $\mathcal{D} \gets \texttt{Shuffle}(\mathcal{D})$
    \EndIf
    
\EndFor
\end{algorithmic}
\end{algorithm}

\subsection{Dual-Branch Decoder}
\label{sec:decoder}

With the proposed bit-compressed tokenization, each $L$-bit message is partitioned into $C$ chunks with $n$ bits per chunk, where each chunk corresponds to one of $2^n$ discrete states. This compact representation improves token efficiency while increasing decoding difficulty due to the enlarged per-token state space. Chunk-level prediction aligns with the tokenizer structure but provides coarse supervision for bit recovery, whereas flat bit prediction offers direct supervision but ignores chunk structure. To leverage both advantages, a dual-branch decoder $\mathcal{D}_M$ is introduced with a shared trainable projection $\boldsymbol{f}'=g_\theta(\boldsymbol{f})\in\mathbb{R}^{512}$, where $\boldsymbol{f}\in\mathbb{R}^{512}$ is the $\ell_2$-normalized CLIP feature. Both branches take $\boldsymbol{f}'$ as input, and their losses jointly update the shared projection. Algorithm~\ref{alg:train_overview} summarizes the training procedure of the dual-branch decoder. \par\vspace{0.8\baselineskip}\noindent

\noindent\textbf{Chunk branch.}
This branch explicitly predicts the $C$ discrete chunks defined by the tokenizer, instead of $L$ flat bits. A two-layer MLP with GELU activation projects the shared feature $\boldsymbol{f}'$ into a sequence of chunk embeddings:
\begin{equation}
  \boldsymbol{H}^{(0)} = \operatorname{Reshape}\bigl(\phi(\boldsymbol{f}')\bigr) + \boldsymbol{P}_s,
  \label{eq:chunk_embed}
\end{equation}
where $\phi$ maps $\mathbb{R}^{512}$ to $\mathbb{R}^{C\times d}$ and $\boldsymbol{P}_s \in \mathbb{R}^{C \times d}$ is a learnable positional embedding. A transformer encoder~\cite{vaswani2017attention} with four attention heads is then applied to model inter-chunk dependencies:
\begin{equation}
  \boldsymbol{H} = [\boldsymbol{h}_1,\ldots,\boldsymbol{h}_C] = \operatorname{TransEnc}\bigl(\boldsymbol{H}^{(0)}\bigr).
\end{equation}
A layer normalization followed by a shared linear classifier produces chunk-level logits:
\begin{equation}
  \boldsymbol{s}_i = \boldsymbol{W}_s \operatorname{LN}(\boldsymbol{h}_i) + \boldsymbol{b}_s \in \mathbb{R}^{2^n}, \quad i=1,\ldots,C.
\end{equation}
The chunk-level loss is defined as 
\begin{equation}
    \mathcal{L}_{\mathrm{chunk}} = \sum\limits_{i=1}^{C}\operatorname{CrossEntropy}(\boldsymbol{s}_i, y_i),
\end{equation} where $y_i$ indicates the ground-truth chunk index.

To bridge chunk-level and bit-level supervision, a fixed binary codebook $\mathbf{B}\in\{0,1\}^{2^n \times n}$ is constructed, where each row encodes the $n$-bit representation consistent with the tokenizer. The predicted bit probabilities for chunk $i$ are obtained by marginalizing over chunk states:
\begin{equation}
  \tilde{p}_{i,r}
  =
  \sum_{j=0}^{2^n-1}
  \operatorname{Softmax}(\boldsymbol{s}_i)_j \, \mathbf{B}[j,r],
  \quad r=1,\ldots,n.
  \label{eq:state_to_bit}
\end{equation}
These are concatenated across chunks to form $\tilde{\boldsymbol{p}}\in[0,1]^L$, with padding positions removed when necessary. The predicted logits are computed as
\begin{equation}
  \tilde{M} = \log \tilde{\boldsymbol{p}} - \log(1-\tilde{\boldsymbol{p}}),
\end{equation}
and we have the projected bit loss via a binary cross-entropy (BCE) loss function as 
\begin{equation}
    \mathcal{L}_{\mathrm{proj}} = \operatorname{BCE}(\tilde{M}, M),
\end{equation}
where $M$ denotes the ground-truth binary message.

\begin{figure}[t]
\centering
\includegraphics[width=1\columnwidth]{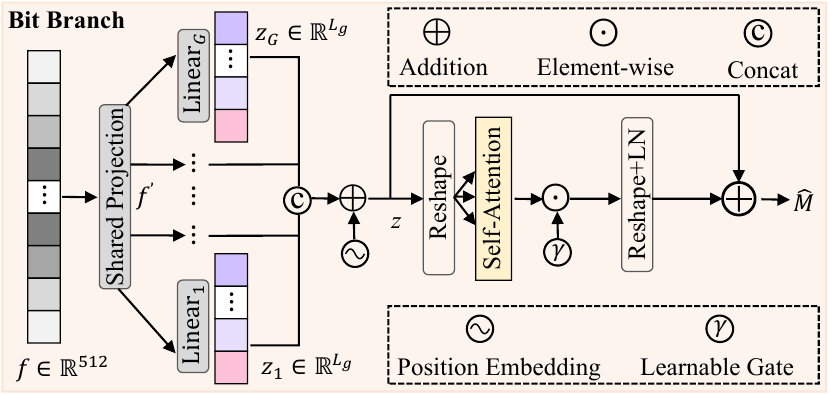} 
\par\vspace{-2mm}
\caption{Illustration of the bit branch in the proposed dual-branch decoder.}
\label{fig_decoder}
\end{figure}

\noindent\textbf{Bit branch.}
While the chunk branch preserves the bit-chunk structure induced by tokenization, watermark embedding also requires direct bit-level supervision for reliable message recovery. To this end, a bit branch is introduced for direct bit prediction, producing bit logits $\hat{M}\in\mathbb{R}^{L}$, as illustrated in Fig.~\ref{fig_decoder}. The shared feature $\boldsymbol{f}'$ is fed into $G$ parallel linear heads to produce group-wise bit logits:
\begin{equation}
  \boldsymbol{z}_i = \varphi_i(\boldsymbol{f}') \in \mathbb{R}^{L_g}, \quad i=1,\ldots,G,
\end{equation}
where $L_g=L/G$. After concatenation, the raw bit logits are
\begin{equation}
  \boldsymbol{z} =
  \operatorname{Concat}(\boldsymbol{z}_1,\ldots,\boldsymbol{z}_G) + P_L
  \in \mathbb{R}^{L}.
\end{equation}
With positional encoding, $\boldsymbol{z}$ is reshaped into a group matrix $\boldsymbol{Z}\in\mathbb{R}^{L_g\times G}$ and refined by a gated self-attention layer:
\begin{equation}
  \boldsymbol{Q}=\boldsymbol{Z}\boldsymbol{W}_Q,\quad
  \boldsymbol{K}=\boldsymbol{Z}\boldsymbol{W}_K,\quad
  \boldsymbol{V}=\boldsymbol{Z}\boldsymbol{W}_V,
\end{equation}
\begin{equation}
  \boldsymbol{Z}' =
  \boldsymbol{\gamma}\odot
\operatorname{Softmax}\!\Bigl(\frac{\boldsymbol{Q}\boldsymbol{K}^{\top}}{\sqrt{G}}\Bigr)\boldsymbol{V},
  \label{eq:gated_attn}
\end{equation}
where $\boldsymbol{\gamma}\in\mathbb{R}^{G}$ is a learnable gate vector and $\odot$ denotes broadcast element-wise multiplication. The final bit logits are obtained as 
\begin{equation}
    \hat{M} = \operatorname{LN} \left( \operatorname{Reshape}(\boldsymbol{Z}^{\prime})\right) + \boldsymbol{z},
\end{equation}
where $\operatorname{Reshape}$ flattens $\mathbb{R}^{L_g \times G}$ into $\mathbb{R}^{L}$. The direct bit-level loss is defined as $\mathcal{L}_{\text{bit}} = \operatorname{BCE}(\hat{M}, M)$.

\subsection{Hard-Message Sampling}
\label{sec:hgdr}

While bit-compressed tokenization and the dual-branch decoder enable high-capacity message representation and recovery, decoder pre-training remains affected by fixed-subset bias in the enlarged message space. As the space grows exponentially with message length, exhaustive training becomes infeasible, leading the decoder to overfit to a limited set of seen messages and reducing generalization to unseen ones. To address this issue, a hard-message sampling strategy is introduced to progressively refresh the training buffer by retaining difficult samples and introducing new messages.

Hard-message sampling maintains two core structures during decoder pre-training: a fixed-size message buffer $\mathcal{D}$ and an accuracy record dictionary $\mathcal{S}$. Initially, $\mathcal{D}$ is populated with $K$ randomly sampled binary messages of length $L$, while $\mathcal{S}$ is empty. After each training iteration, $\mathcal{S}$ is updated using online decoding feedback. Given a ground-truth message $M$ and its prediction $\hat{M}$, messages with at least one incorrect bit are identified. For each such message $M_k$, its integer representation $m_k$ is computed and its bit-wise accuracy $\text{acc}_k$ is calculated. If $m_k$ is not in $\mathcal{S}$, it is added with value $\text{acc}_k$; otherwise, its record is updated via exponential smoothing:
\begin{equation}
\mathcal{S}[m_k] \leftarrow
\begin{cases}
\text{acc}_k, & \text{if } m_k \notin \mathcal{S}, \\
0.5 \cdot \mathcal{S}[m_k] + 0.5 \cdot \text{acc}_k, & \text{otherwise}.
\end{cases}
\end{equation}
This mechanism prioritizes persistently difficult messages while avoiding unnecessary updates for easily decoded ones.

At the end of each epoch $e$, the buffer $\mathcal{D}$ is refreshed by resampling from the hard-message set in $\mathcal{S}$ and the unseen message space. Before the predefined freeze epoch $E_{\text{freeze}}$, the proportion of hard messages is increased linearly:
\begin{equation}
r_{\text{hard}} = \min(1.0, \tau_0 + \alpha \cdot e),
\end{equation}
where $\tau_0$ is the initial hard ratio and $\alpha$ is a slope parameter, and $r_{\text{hard}}$ is the proportion of hard messages to retain. To construct the hard message subset, messages in $\mathcal{S}$ with historical accuracy below a threshold $\sigma$ are selected, sorted in ascending order of accuracy, and the bottom $\lfloor r_{\mathrm{hard}} \cdot K \rfloor$ samples are retained to form the hard subset $\mathcal{H}$. To maintain diversity, the remaining $K-|\mathcal{H}|$ binary messages are sampled uniformly without replacement from $\{0,1\}^L\setminus\mathcal{H}$ to form $\mathcal{N}$. The buffer is then updated as
\begin{equation}
\mathcal{D} \leftarrow \texttt{Shuffle}(\mathcal{H}\cup\mathcal{N}).
\end{equation}
After training epoch $E_{\text{freeze}}$, the bottom-$K$ messages in $\mathcal{S}$ ranked by historical accuracy are selected once to form the fixed training buffer. The training pool then remains fixed, with only its order shuffled in subsequent epochs. By updating the buffer across epochs rather than fixing it at initialization, hard-message sampling progressively expands message-space coverage during decoder pre-training. This mitigates fixed-subset bias and improves the recovery of unseen messages.

\section{Experiments}
\label{sec:experiments}

In this section, we evaluate the proposed BitC-3DGS across both synthetic objects and real-world scenes to assess its capacity, visual fidelity, and robustness. We compare against representative 3D Gaussian Splatting watermarking methods under varying message lengths and distortion settings, and further analyze performance in both seen and unseen message regimes. In addition, we provide extensive ablation and sensitivity studies to validate the effectiveness of each component in our framework.

\begin{table*}[htb]
\setlength{\tabcolsep}{0.9mm}
    \centering
    \caption{\textbf{Comparisons of the State-of-the-art Methods} on Blender and LLFF Datasets for Bit-Accuracy and Rendering Qualities. We Show the Results in 32, 48, and 64 Bits. The Best and Second-best Results are Highlighted in \textbf{Bold} and \underline{Underline} Formats.}
    \renewcommand{\arraystretch}{1.1} 
    \resizebox{1.0\linewidth}{!}
    {
    \begin{tabular}{l | ccc ccc ccc}
        \toprule
        \multirow[c]{2}{*}{Methods} &  \multicolumn{3}{c}{32 bits} & \multicolumn{3}{c}{48 bits} & \multicolumn{3}{c}{64 bits} \\
        \cmidrule(lr){2-4} \cmidrule(lr){5-7} \cmidrule(lr){8-10}
        & Bit-Acc $\uparrow$ & PSNR / SSIM $\uparrow$ & LPIPS $\downarrow$ & Bit-Acc $\uparrow$ & PSNR / SSIM $\uparrow$ & LPIPS $\downarrow$ & Bit-Acc $\uparrow$ & PSNR / SSIM $\uparrow$ & LPIPS $\downarrow$ \\
        \midrule
        GaussianMarker \cite{gaussianmarker} & 98.85 & 33.98 / 0.979 & 0.0163 & 98.25 & 32.12 / \underline{0.972} & 0.0234 & 91.69 & 28.44 / 0.923 & 0.0497 \\     
        3DGS \cite{3dgs} + WateRF \cite{waterf} & 93.42 & 30.49 / 0.956 & 0.0500 & 84.16 & 29.92 / 0.951 & 0.0530 & 75.10 & 25.81 / 0.883 & 0.1080 \\
        3DGS \cite{3dgs} + StegaNeRF \cite{steganerf} & 93.15 & 32.68 / 0.953 & 0.0490 & 89.43 & 32.72 / 0.954 & 0.0480 & 85.27 & 30.66 / 0.925 & 0.0920 \\
        3D-GSW \cite{3d-gsw} & 97.37 & 35.08 / 0.978 & 0.0430 & 93.72 & 33.31 / 0.970 & 0.0450 & 90.45 & 32.47 / 0.967 & 0.0490 \\
        GuardSplat \cite{guardsplat}~\textit{[Random]} & 98.69 & 39.31 / \textbf{0.995} & 0.0027 & 96.41 & 37.86 / \textbf{0.992} & 0.0042 & 92.93 & 35.95 / 0.984 & 0.0070 \\
        GuardSplat \cite{guardsplat}~\textit{[In]} & 99.04 & 39.40 / \underline{0.994} & \textbf{0.0022} & 98.29 & \underline{38.90} / \textbf{0.992} & \textbf{0.0028} & 95.07 & 36.05 / 0.987 & 0.0065 \\
        \midrule
        \rowcolor{black!10} BitC-3DGS ~\textit{[Random]} & \underline{99.67} & \underline{40.35} / \textbf{0.995} & \underline{0.0023} & \underline{98.87} & 38.51 / \textbf{0.992} & \underline{0.0032} & \underline{97.35} & \underline{36.30} / \underline{0.988} & \underline{0.0051} \\
        \rowcolor{black!10} BitC-3DGS ~\textit{[In]} & \textbf{99.71} & \textbf{40.73} / \textbf{0.995} & \textbf{0.0022} & \textbf{99.17} & \textbf{39.10} / \textbf{0.992} & \textbf{0.0028} & \textbf{97.90} & \textbf{36.60} / \textbf{0.989} & \textbf{0.0047} \\
        \bottomrule
    \end{tabular}
    }
    \label{tab:experiment_result}
\end{table*}

\subsection{Experimental Settings}

\noindent\textbf{Datasets.} We use two benchmark datasets for evaluation: the Blender dataset~\cite{nerf}, consisting of eight synthetic objects with controlled camera trajectories, and the LLFF dataset~\cite{llff}, comprising seven real-world forward-facing scenes captured with handheld cameras. Blender provides clean renderings with accurate geometry and view-dependent effects for evaluation under ideal conditions, while LLFF contains real scenes with natural illumination, imperfect geometry, and larger view variations, enabling robustness assessment in practical settings. Following GuardSplat~\cite{guardsplat}, we use orbital trajectories for Blender and spiral trajectories for LLFF. For each scene, we randomly sample 200 views per epoch for watermark embedding and evaluate on 200 fixed views sampled at uniform parameter intervals along the same trajectory. Bit accuracy and fidelity metrics are averaged over test views for quantitative evaluation, while bit accuracy in Figs.~\ref{fig_vis}--\ref{fig_vis_ablation} is reported for the displayed views. Fidelity is measured against original 3DGS renderings at matching viewpoints. \par\vspace{0.8\baselineskip}\noindent

\noindent\textbf{Metrics.} We evaluate BitC-3DGS for high-capacity  watermarking along four dimensions.
(1)~\emph{Capacity}: We measure the bit accuracy for message lengths $L\in\{16, 32, 48, 64, 96, 128\}$. 
(2)~\emph{Fidelity:} We assess visual quality using standard perceptual metrics: PSNR, SSIM~\cite{ssim}, and LPIPS~\cite{lpips} between watermarked and original 3DGS renderings at the same viewpoints.
(3)~\emph{Robustness:} To test resilience against both 2D visual perturbations and 3D model distortions, we report extraction accuracy under seven individual 2D distortions, one composite 2D attack on rendered images, and three model-space perturbations on the Gaussian representation.
(4)~\emph{Seen--Unseen Recovery:} We report bit accuracy for three settings:
\begin{itemize}[leftmargin=*]
\item \textit{[In]}: Messages sampled from the final training buffer $\mathcal{D}$ (of size $\min(2^{L},K)$).
\item \textit{[Out]}: Messages sampled from
$\{0,1\}^L \setminus \mathcal{D}$, outside the final decoder-training buffer.
\item \textit{[Random]}: The equally weighted average of \textit{[In]} and \textit{[Out]}, reflecting recovery when messages are sampled from both the training buffer $\mathcal{D}$ and the unseen space $\{0,1\}^L \setminus \mathcal{D}$. \par\vspace{0.8\baselineskip}\noindent
\end{itemize}

\noindent\textbf{Baselines.} We compare BitC-3DGS with representative 3DGS watermarking baselines, including GaussianMarker~\cite{gaussianmarker}, 3D-GSW~\cite{3d-gsw}, 3DGS+WateRF~\cite{waterf}, 3DGS+StegaNeRF~\cite{steganerf}, GuardSplat~\cite{guardsplat}, and a strengthened GuardSplat variant with our bit-compressed tokenizer (GuardSplat \textit{w/} $\texttt{Tokenizer})$. For GuardSplat, we evaluate under two settings: \textbf{\textit{[In]}}, where messages are sampled from its fixed decoder training set, and \textbf{\textit{[Random]}}, which reports the equally weighted average of its \textit{[In]} and \textit{[Out]} results under the same evaluation protocol used for BitC-3DGS. For baselines without an explicit decoder training buffer, we follow their respective evaluation protocols and use a randomly sampled message for watermark embedding and testing.

\subsection{Implementation Details}
\label{sec:implementation_details}

All training and evaluation are conducted on a single NVIDIA RTX 4090 GPU.

\textit{For pre-trained 3DGS.}
We adopt the official PyTorch implementation of 3DGS~\cite{3dgs}. For each object in Blender~\cite{nerf} and each scene in LLFF~\cite{llff}, we train a standard 3DGS model using the default settings of the official repository.

\textit{For message decoder.}
We use frozen CLIP~\cite{clip} ViT-B/32 text and image encoders with 512-dimensional outputs. The dual-branch decoder is trained using Adam~\cite{adam} (learning rate $5\times10^{-3}$, weight decay $10^{-6}$, batch size 64) with a buffer of $K=2048$ binary messages. For $L\in\{16,32,48\}$ and $L\in\{64,96,128\}$, training lasts 150 and 300 epochs, with $\mathcal{D}$ frozen after 100 and 200 epochs, respectively. The corresponding initial hard ratios and warm-up slopes are $(\tau_0,\alpha)=(0.30,0.0045)$ and $(0.25,0.0025)$. Before freezing, hard-message candidates are selected using the historical accuracy threshold $\sigma=0.98$. We use $(n,G)=(1,1)$ for $L=16$, $(2,4)$ for $L\in\{32,48,64,96\}$, and $(4,4)$ for $L=128$, with decoder loss weights $(\lambda_s,\lambda_p,\lambda_b)=(1.0,0.25,1.0)$. After pre-training, the decoder is frozen, and one checkpoint per payload configuration is reused across all scenes and datasets.

\textit{For watermarking 3DGS models.}
We adopt SH-offset optimization~\cite{guardsplat} using Adam with learning rates of $5\times10^{-3}$ and $2.5\times10^{-4}$ for the DC and remaining SH offsets, respectively, $(\beta_1,\beta_2)=(0.9,0.999)$, $\epsilon=10^{-15}$, and a batch size of 24. Optimization runs for 150 epochs ($L\le32$), 200 epochs ($L\in\{48,64\}$), and 300 epochs ($L\ge96$). The frozen decoder's bit branch provides message supervision, while image fidelity is maintained relative to renderings of the original, unwatermarked model at the same viewpoints. We set $\lambda_{\mathrm{ssim}}=0.2$, $\lambda_{\mathrm{image}}=1$, $\lambda_{\mathrm{bit}}=0.03$, and $\lambda_{\mathrm{off}}=10$.

\textit{For distortion.}
We apply per-view stochastic 2D distortions before message decoding during watermark embedding. The image-space distortion settings are based on GuardSplat~\cite{guardsplat}, while the model-space perturbations used for robustness evaluation follow 3D-GSW~\cite{3d-gsw}.
\textit{Image-space:} Gaussian noise ($\mu=0$, $\sigma=0.01$), rotation with angles uniformly sampled from $[-\pi/6,\pi/6]$, scaling (75\% of the original), Gaussian blur ($\sigma=0.1$), center cropping (40\% of the original), brightness jitter with multiplicative factors uniformly sampled from $[0.5,1.5]$, JPEG compression with a quality factor of 50, and a composite crop--blur--JPEG pipeline.
\textit{Model-space:} randomly removing 20\% of Gaussians, randomly cloning 20\% of Gaussians, and adding Gaussian noise ($\mu=0$, $\sigma=0.1$) to SH coefficients.

\begin{table*}[htb]
\setlength{\tabcolsep}{0.9mm}
    \centering
    \caption{Comparisons with Baselines \cite{gaussianmarker, guardsplat} on Blender and LLFF Datasets in High-Capacity Settings.}
    \renewcommand{\arraystretch}{1.1} 
    \begin{tabular}{l | cccc cccc}
        \toprule
        \multirow[c]{2}{*}{Methods} &  \multicolumn{4}{c}{96 bits} & \multicolumn{4}{c}{128 bits}\\
        \cmidrule(lr){2-5} \cmidrule(lr){6-9}
        & Bit-Acc $\uparrow$ & PSNR $\uparrow$ & SSIM $\uparrow$ & LPIPS $\downarrow$ & Bit-Acc $\uparrow$ & PSNR $\uparrow$ & SSIM $\uparrow$ & LPIPS $\downarrow$ \\
        \midrule
        GaussianMarker \cite{gaussianmarker} & 76.69 & 28.98 & 0.932 & 0.0484 & -- & -- & -- & -- \\
        GuardSplat \cite{guardsplat} \textit{w/} $\texttt{Tokenizer}$~\textit{[Random]} & 88.41 & 34.48 & 0.982 & 0.0089 & 80.84 & 33.05 & 0.976 & 0.0125 \\
        GuardSplat \cite{guardsplat} \textit{w/} $\texttt{Tokenizer}$~\textit{[In]} & 90.57 & \underline{34.84} & \underline{0.984} & 0.0074 & 84.45 & 33.01 & 0.976 & 0.0099 \\
        \midrule
        \rowcolor{black!10} BitC-3DGS ~\textit{[Random]} & \underline{93.97} & 34.59 & \underline{0.984} & \underline{0.0072} & \underline{90.59} & \underline{33.69} & \underline{0.980} & \underline{0.0082} \\
        \rowcolor{black!10} BitC-3DGS ~\textit{[In]} & \textbf{94.27} & \textbf{35.11} & \textbf{0.986} & \textbf{0.0066} & \textbf{91.42} & \textbf{34.32} & \textbf{0.983}  & \textbf{0.0068} \\                                       
        \bottomrule
    \end{tabular}
    \label{tab:capacity_comparision}
\end{table*}

\begin{table*}[!t]
\centering
\caption{Comparison of State-of-the-Art Methods on Blender and LLFF Under Visual Distortions.
}
\label{table:robustness}
\setlength{\tabcolsep}{1mm}
\begin{adjustbox}{max width=\textwidth,center}
{
\begin{tabular}{l| ccccccccc}
\toprule
\multirow{2}{*}{Methods}               &\multirow{2}{*}{None} &Noise   &Rotation   &Scaling   &Blur      &Crop   &Brightness  &JPEG     &Combined\\
&           &($\sigma=0.01$)&($\pm$$\pi$/6)&(75\%)&($\sigma$=$0.1$)&(40\%)  &(0.5$\sim$1.5)&(50\% quality) &(Crop, Blur, JPEG)             \\\midrule
\rowcolor{gray!5} \multicolumn{10}{@{}l}{\textit{32 bits}} \\
GaussianMarker \cite{gaussianmarker} &98.85 &97.20 &80.72 &98.03  &93.57  &91.14  &94.97   &90.02    &89.10  \\
3D-GSW \cite{3d-gsw} &97.37 &83.84 &87.94 &94.64  &96.01  &92.86  &92.41   &92.65    &90.84  \\
GuardSplat \cite{guardsplat}~\textit{[In]}  & 99.04 & 97.02 & 92.83  & 97.45 & 96.10  & 93.77 & 96.86  & 92.57  & 91.86   \\
\rowcolor{black!10} BitC-3DGS~\textit{[In]}  & \textbf{99.71}  & \textbf{97.80} & \textbf{93.21} & \textbf{98.32} & \textbf{97.01} & \textbf{94.55} & \textbf{97.34} & \textbf{93.05} & \textbf{92.14} \\
\midrule
\rowcolor{gray!5} \multicolumn{10}{@{}l}{\textit{96 bits}} \\
GuardSplat \cite{guardsplat} \textit{w/} $\texttt{Tokenizer}$~\textit{[In]}  & 90.57 & 87.65 & 83.76  & 88.98 & 87.43  & 84.99 & 88.42  & 83.31  & 82.39   \\
\rowcolor{black!10} BitC-3DGS~\textit{[In]}  & \textbf{94.27}  & \textbf{91.76} & \textbf{88.79} & \textbf{93.58} & \textbf{91.57} & \textbf{88.11} & \textbf{91.93} & \textbf{87.63} & \textbf{88.02} \\
\bottomrule
\end{tabular}
}
\end{adjustbox}
\end{table*}

\subsection{Quantitative Results}

\noindent \textbf{Capacity and Fidelity.}
Tables~\ref{tab:experiment_result} and~\ref{tab:capacity_comparision} report results under 32–128 bit payloads. At 32--64 bits, BitC-3DGS consistently achieves the highest bit accuracy under both \textit{[In]} and \textit{[Random]} settings, with the gap widening as capacity increases. For instance, at 64 bits, BitC-3DGS reaches 97.35\% and 97.90\% bit accuracy under \textit{[Random]} and \textit{[In]}, compared to 92.93\% and 95.07\% for GuardSplat. This demonstrates that bit-compressed tokenization and the dual-branch decoder improve reliability at higher payloads.

Table~\ref{tab:capacity_comparision} further evaluates the 96-bit and 128-bit settings. To test whether compact tokenization alone is sufficient, we equip GuardSplat~\cite{guardsplat} with our bit-compressed tokenization as a strong baseline. BitC-3DGS still consistently outperforms it, indicating that effective high-capacity recovery requires both compact tokenization and a dedicated decoder. At 96 bits, BitC-3DGS achieves 93.97\% and 94.27\% bit accuracy under \textit{[Random]} and \textit{[In]}, compared to 88.41\% and 90.57\% for GuardSplat w/ \texttt{Tokenizer}. At 128 bits, the gap widens further, with BitC-3DGS reaching 90.59\% and 91.42\%, versus 80.84\% and 84.45\%. These gains are achieved without sacrificing rendering quality, with consistently high PSNR/SSIM and low LPIPS across settings; for example, at 128 bits (\textit{[In]}), BitC-3DGS improves PSNR from 33.01 to 34.32 and reduces LPIPS from 0.0099 to 0.0068. Overall, BitC-3DGS achieves reliable, high-capacity watermark recovery within the fixed token budget while preserving high-fidelity 3DGS rendering.

\begin{table}[htb]
\setlength{\tabcolsep}{0.9mm}
    \centering
    \caption{Comparison with Baseline \cite{guardsplat} on Blender and LLFF Across Message Lengths.}
    \renewcommand{\arraystretch}{1.1} 
    \resizebox{1.0\linewidth}{!}
    {
    \begin{tabular}{l | ccc ccc}
        \toprule
        \multirow[c]{2}{*}{Bit Lengths} &  \multicolumn{3}{c}{BitC-3DGS} & \multicolumn{3}{c}{Baseline \cite{guardsplat}} \\
        \cmidrule(lr){2-4} \cmidrule(lr){5-7}
        & Bit-Acc $\uparrow$ & PSNR $\uparrow$ & SSIM $\uparrow$ & Bit-Acc $\uparrow$ & PSNR $\uparrow$ & SSIM $\uparrow$  \\
        \midrule
        32 bits \textit{[In]}  & 99.71 $\uparrow_{0.67}$ & 40.73 & 0.995 & 99.04 & 39.40 & 0.994 \\      
        32 bits \textit{[Out]} & 99.63 $\uparrow_{1.29}$ & 39.97 & 0.994 & 98.34 & 39.22 & 0.996 \\ 
        \midrule
        48 bits \textit{[In]}  & 99.17 $\uparrow_{0.88}$ & 39.10 & 0.992 & 98.29 & 38.90 & 0.992 \\      
        48 bits \textit{[Out]} & 98.57 $\uparrow_{4.05}$ & 37.92 & 0.992 & 94.52 & 36.82 & 0.991 \\ 
        \midrule
        64 bits \textit{[In]} & 97.90 $\uparrow_{2.83}$ & 36.60 & 0.989  & 95.07 & 36.05 & 0.987  \\    
        64 bits \textit{[Out]} & 96.80 $\uparrow_{6.02}$ & 36.00 & 0.987 & 90.78 & 35.85 & 0.981 \\ 
        \bottomrule
    \end{tabular}
    }
    \label{tab:guardsplat_comparision}
\end{table}

\noindent \textbf{Seen--Unseen Recovery.}
Table~\ref{tab:guardsplat_comparision} compares recovery on seen (\textit{[In]}) and unseen (\textit{[Out]}) messages. Across all capacities, BitC-3DGS consistently outperforms the baseline while maintaining comparable or better rendering fidelity. It also exhibits a much smaller \textit{[In]}/\textit{[Out]} gap that remains stable as capacity increases: 0.08 (BitC-3DGS) vs. 0.70 (baseline) at 32 bits, 0.60 vs. 3.77 at 48 bits, and 1.10 vs. 4.29 at 64 bits.
Performance on unseen messages improves more clearly at higher payloads, where BitC-3DGS surpasses the baseline by 4.05 points at 48 bits and 6.02 points at 64 bits in bit accuracy. These results indicate that BitC-3DGS reduces fixed-subset bias and enables more reliable recovery for unseen messages as the message space expands.\par\vspace{0.8\baselineskip}\noindent

\noindent \textbf{Robustness.}
We evaluate robustness under 2D visual distortions and 3D model perturbations at 32-bit and 96-bit payloads, using \textit{[In]} messages for BitC-3DGS and the GuardSplat variants.
\textbf{1) 2D visual distortions:}
Table~\ref{table:robustness} shows that BitC-3DGS achieves the highest bit accuracy across all tested distortions at 32 bits. At 96 bits, it outperforms GuardSplat \textit{w/} \texttt{Tokenizer} by 3.12--5.63 percentage points across the evaluated attacks. Under the challenging \textit{Combined} attack (Crop + Blur + JPEG), it achieves 88.02\% accuracy, compared with 82.39\% for the baseline.
\textbf{2) 3D model perturbations:}
As shown in Table~\ref{tab:robustness_3d}, BitC-3DGS remains competitive at 32 bits, achieving the highest accuracy under pruning (97.51\%). At 96 bits, it achieves 86.03\%, 94.18\%, and 92.93\% accuracy under SH noise, pruning, and cloning, respectively, exceeding GuardSplat \textit{w/} \texttt{Tokenizer} by 4.93, 5.97, and 5.17 percentage points. These results demonstrate that the recovery advantage persists at higher capacity under both image-space and model-space perturbations.

\begin{figure*}[t]\centering
\includegraphics[width=1 \textwidth]{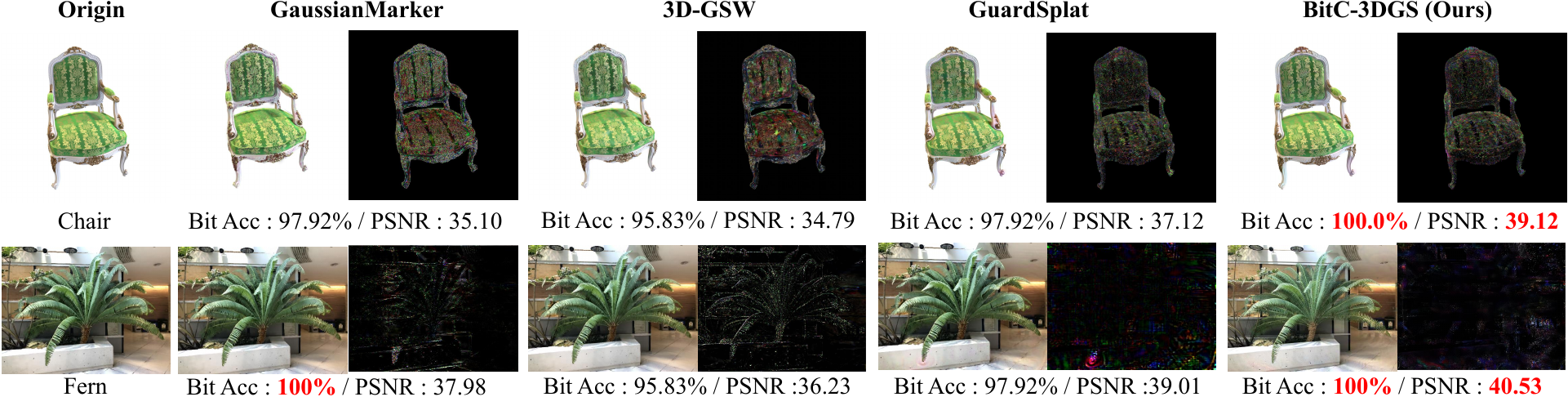}
\par\vspace{-2mm}
\caption{\label{fig_vis} Visual comparisons with baselines \cite{gaussianmarker, 3d-gsw, guardsplat} under $L=48$ bits on the Blender \cite{nerf} dataset and LLFF \cite{llff} dataset.  For each setting, the left image shows the rendered view, and the right shows the difference ($\times5$) map relative to the original 3DGS renderings.} 
\end{figure*}

\begin{table}[!ht]
\setlength{\tabcolsep}{0.9mm}
    \centering
    \caption{Robustness Under Model Distortion. We Show the Results on 32 and 96 bits.    The Best and Second-best Results are Highlighted in \textbf{Bold} and \underline{Underline} Formats.}
    \renewcommand{\arraystretch}{1.1} 
    \resizebox{1.0\linewidth}{!}
    {
    \begin{tabular}{l | cccc}
        \toprule
        &  \multicolumn{4}{c}{Bit-Accuracy(\%) $\uparrow$} \\
        \cmidrule(lr){2-5}
        Methods & \begin{tabular}[c]{@{}c@{}}No \\ Distortion \end{tabular} & \begin{tabular}[c]{@{}c@{}} Gaussian Noise \\ ($\sigma$ = 0.1) \end{tabular} & \begin{tabular}[c]{@{}c@{}} Removing \\ (20 \%) \end{tabular} & \begin{tabular}[c]{@{}c@{}} Cloning \\ (20 \%) \end{tabular} \\
        \midrule
        \rowcolor{gray!5} \multicolumn{5}{@{}l}{\textit{32 bits}} \\
        GaussianMarker \cite{gaussianmarker} &98.85 & \textbf{92.81}  & 96.81  & \textbf{97.14} \\
        3D-GSW \cite{3d-gsw} & 97.37 & 89.11 & \underline{96.87} & 95.99 \\
        GuardSplat \cite{guardsplat}~\textit{[In]} &\underline{99.04}  & 89.73  & 96.31 & 95.70 \\
        \rowcolor{black!10} BitC-3DGS ~\textit{[In]} & \textbf{99.71} & \underline{92.28} & \textbf{97.51}  & \underline{96.39}  \\
        \midrule
        \rowcolor{gray!5} \multicolumn{5}{@{}l}{\textit{96 bits}} \\
        GuardSplat \textit{w/} $\texttt{Tokenizer}$~\textit{[In]}  & 90.57 & 81.10 & 88.21  & 87.76  \\
        \rowcolor{black!10} BitC-3DGS ~\textit{[In]} & \textbf{94.27} & \textbf{86.03} & \textbf{94.18}  & \textbf{92.93}  \\
        \bottomrule
    \end{tabular}
    }
    \label{tab:robustness_3d}
\end{table}

\begin{figure}[!t]
\centering
\includegraphics[width=1.0\linewidth]{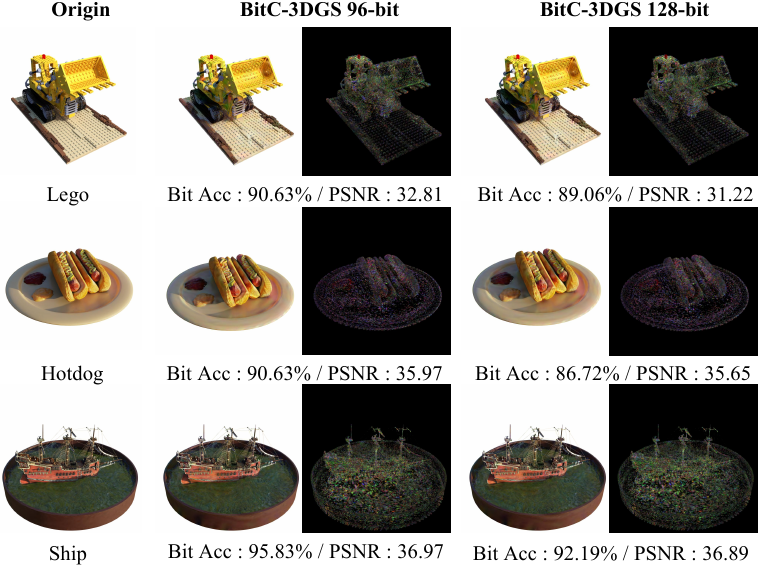} 
\par\vspace{-2mm}
\caption{Qualitative results of BitC-3DGS under 96-bit and 128-bit payloads on the Blender \cite{nerf} dataset. For each setting, the left image shows the rendered view, and the right shows the difference ($\times5$) map relative to the origin.}
\label{fig_blender_highcap}
\end{figure}

\begin{figure}[!t]
\centering
\includegraphics[width=0.99\linewidth]{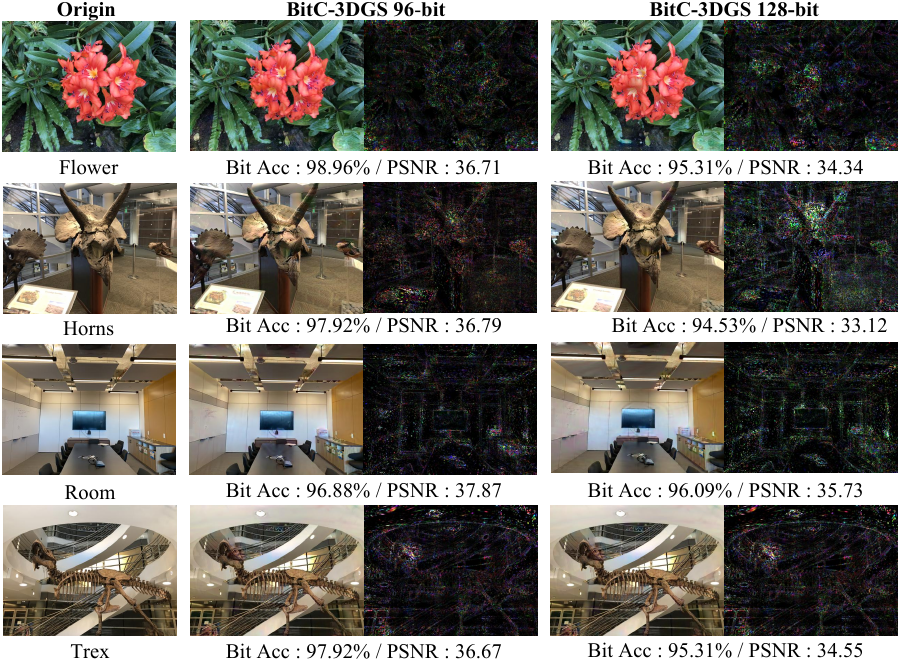} 
\par\vspace{-2mm}
\caption{Qualitative results of BitC-3DGS under 96-bit and 128-bit payloads on the LLFF \cite{llff} dataset. For each setting, the left image shows the rendered view, and the right shows the difference ($\times5$) map relative to the origin.}
\label{fig_llff_highcap}
\end{figure}

\subsection{Qualitative Results}
\noindent \textbf{Low-Capacity.}
Fig.~\ref{fig_vis} compares the visual fidelity of BitC-3DGS with representative baselines under a 48-bit payload. For each method, we show renderings and residual maps relative to the original views. Baseline methods introduce noticeably stronger distortions in the residuals: GaussianMarker~\cite{gaussianmarker} and 3D-GSW~\cite{3d-gsw} produce scattered high-frequency artifacts, while GuardSplat~\cite{guardsplat} leads to more structured perturbations, particularly along object boundaries and textured regions. In contrast, BitC-3DGS generates cleaner residual maps and renderings that are visually closer to the original 3DGS renderings. These qualitative observations are consistent with the quantitative results, showing that BitC-3DGS achieves higher accuracy while maintaining lower distortion. \par\vspace{0.8\baselineskip}\noindent

\noindent \textbf{High-Capacity.} Figs.~\ref{fig_blender_highcap} and~\ref{fig_llff_highcap} present qualitative results of BitC-3DGS under 96-bit and 128-bit payloads on the Blender and LLFF datasets, respectively. Even at these high capacities, the rendered views remain visually close to the original 3DGS renderings, and the corresponding difference maps do not exhibit strong localized artifacts. Although perturbations slightly increase from 96 to 128 bits, overall visual quality remains stable across both datasets, consistent with the quantitative results in terms of high recovery accuracy and PSNR. On Blender objects (\eg, \textit{Lego}, \textit{Hotdog}, and \textit{Ship}), BitC-3DGS preserves global structure and appearance at both capacities. Similarly, on LLFF scenes (\eg, \textit{Flower}, \textit{Horns}, \textit{Room}, and \textit{Trex}), the renderings remain perceptually faithful, with difference maps showing mainly low-amplitude, spatially distributed perturbations rather than localized distortions. Overall, these results demonstrate that BitC-3DGS supports 96-bit and 128-bit payloads with minimal visual degradation on both synthetic and real-world scenes.

\begin{table*}[t]
\setlength{\tabcolsep}{3.2mm}
\renewcommand{\arraystretch}{1.12}
\centering
\caption{Ablation Study of The Two Decoder Branches Under a 64-Bit Setting on the Blender and LLFF Datasets.}
\resizebox{0.7\linewidth}{!}{
\begin{tabular}{c|c| ccc | ccc}
\toprule
\multirow{2}{*}{Setting} 
& \multirow{2}{*}{\begin{tabular}[c]{@{}c@{}}Chunk Branch\end{tabular}}
& \multicolumn{3}{c|}{Bit Branch}
& \multicolumn{3}{c}{64-bit \textit{[In]}} \\
\cmidrule(lr){3-5} \cmidrule(lr){6-8}
& & None & MLP & Ours  & Bit-Acc & PSNR & LPIPS \\
\midrule
Baseline &  &  & \checkmark &  & 95.13 & 35.99 & 0.0066 \\
Setting4 & \checkmark & \checkmark &  &  & 80.35 & \textbf{42.06} & \textbf{0.0015} \\
Setting5 & \checkmark &  & \checkmark &  & 96.02 & 36.10 & 0.0056 \\
Setting6 &  &  &  & \checkmark & \underline{97.79} & 35.20 & 0.0077 \\
\midrule
\rowcolor{black!10} BitC-3DGS & \checkmark &  &  & \checkmark & \textbf{98.10} & \underline{36.61} & \underline{0.0047} \\
\bottomrule
\end{tabular}
}
\label{tab:ablation_decoder}
\end{table*}

\begin{table}[t]
    \centering
    \caption{Ablation Studies of the Three Proposed Modules Under a 64-Bit Setting on the Blender and LLFF Datasets.}
    \renewcommand{\arraystretch}{1.1} 
    \resizebox{1.0\linewidth}{!}
    {
    \begin{tabular}{l| ccc | cc|cc}
        \toprule
        \multirow{2}{*}{Setting} & \multicolumn{3}{c|}{Module} & \multicolumn{2}{c|}{\textit{[In]}} & \multicolumn{2}{c}{\textit{[Out]}} \\
        \cmidrule(lr){2-4} \cmidrule(lr){5-6} \cmidrule(lr){7-8}
        & BCT & DBD & HMS & Bit-Acc & PSNR & Bit-Acc & PSNR \\
        \midrule
        Baseline &  &  &  & 95.07 & 36.05 & 90.78 & 35.85 \\
        Setting1 & \checkmark &  &  & 95.13 & 35.99 & 90.85 & 35.85 \\
        Setting2 & \checkmark & \checkmark & & \textbf{98.10} & \textbf{36.61} & 92.87 & \underline{35.91}  \\
        Setting3 &  &  & \checkmark & 94.99 & 36.03 & \underline{93.76} & 35.90 \\
        \midrule
        \rowcolor{black!10} BitC-3DGS & \checkmark  & \checkmark & \checkmark &\underline{97.90} & \underline{36.60} & \textbf{96.80} & \textbf{36.00}  \\
        \bottomrule
    \end{tabular}
    }
    \label{tab:ablation_study}
\end{table}

\subsection{Ablation Studies}

To better understand BitC-3DGS, we perform ablation studies on compressed tokenization, decoder design, and hard-message sampling. For simplicity, we refer to BitC-3DGS's key components as follows: bit-compressed tokenization (BCT), dual-branch decoder (DBD), and hard-message sampling (HMS). \par\vspace{0.8\baselineskip}\noindent

Table~\ref{tab:ablation_study} shows that the three components address different bottlenecks in high-capacity 3DGS watermarking. Using BCT alone yields only marginal gains, suggesting that relaxing the token budget constraint is insufficient for reliable recovery. Combining it with DBD (Setting2) significantly improves \textit{[In]} accuracy and fidelity, showing that compact tokenization must be paired with effective decoding, but it still leaves a notable \textit{[In]}/\textit{[Out]} gap. HMS (Setting3) mainly reduces this gap by mitigating fixed-subset bias, improving \textit{[Out]} accuracy from 90.78\% to 93.76\% while leaving \textit{[In]} nearly unchanged and reducing the gap from 4.29 to 1.23 points. With all components combined, BitC-3DGS achieves the best balance, reaching 97.90\% (\textit{[In]}) and 96.80\% (\textit{[Out]}) bit accuracy, along with the highest PSNR in the \textit{[Out]} setting. Overall, BCT+DBD ensures high-capacity recovery, while HMS improves generalization to unseen messages, and both are required for optimal performance. \par\vspace{0.8\baselineskip}\noindent

\begin{figure}[t]
\centering
\includegraphics[width=\linewidth]{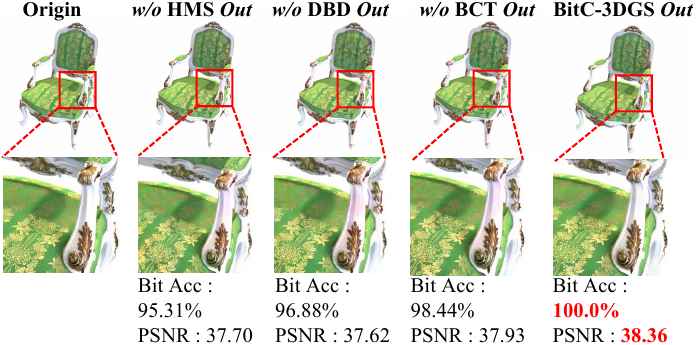} 
\par\vspace{-2mm}
\caption{\textbf{Visual ablation study of the three proposed components} on the chair scene ($L=64$ bits) from the Blender dataset. Each column shows the rendered result and a magnified patch for a specific configuration.}
\label{fig_vis_ablation}
\end{figure}

Table~\ref{tab:ablation_decoder} isolates the roles of the decoder branches at $L=64$ bits. Hard-message sampling is disabled and all settings use \textit{[In]} messages to focus on decoder design. The baseline adopts the original GuardSplat MLP decoder. Setting4 retains only the shared projection and chunk branch, trained with $\lambda_s\mathcal{L}_{\mathrm{chunk}}+\lambda_p\mathcal{L}_{\mathrm{proj}}$. Setting5 combines the chunk branch with the original MLP bit decoder, Setting6 keeps only the proposed bit branch, and BitC-3DGS uses both branches. Results show that a single branch is insufficient. Setting4 achieves very high fidelity (42.06 PSNR, 0.0015 LPIPS) but low bit accuracy (80.35\%), indicating coarse supervision. Setting6 attains strong accuracy (97.79\%) but with reduced fidelity, suggesting stronger visual perturbations. Setting5 improves over the baseline yet remains below BitC-3DGS. The full decoder achieves the best balance, reaching 98.10\% bit accuracy with improved fidelity (36.61 PSNR, 0.0047 LPIPS), confirming that the chunk branch preserves bit-group structure while the bit branch provides direct bit-level supervision. \par\vspace{0.8\baselineskip}\noindent

\noindent \textbf{Visualization.}
Fig.~\ref{fig_vis_ablation} presents a qualitative comparison on the Chair scene under the \textit{[Out]} setting. Removing any module reduces unseen-message recovery or introduces stronger local artifacts. Removing HMS notably degrades \textit{[Out]} performance, while removing DBD increases local distortions, highlighting the need for effective decoding after BCT. The full BitC-3DGS model achieves the best bit accuracy and PSNR while preserving the cleanest local structure, consistent with the quantitative ablations.

\begin{figure}[!t]
\centering
\includegraphics[width=0.8\linewidth]{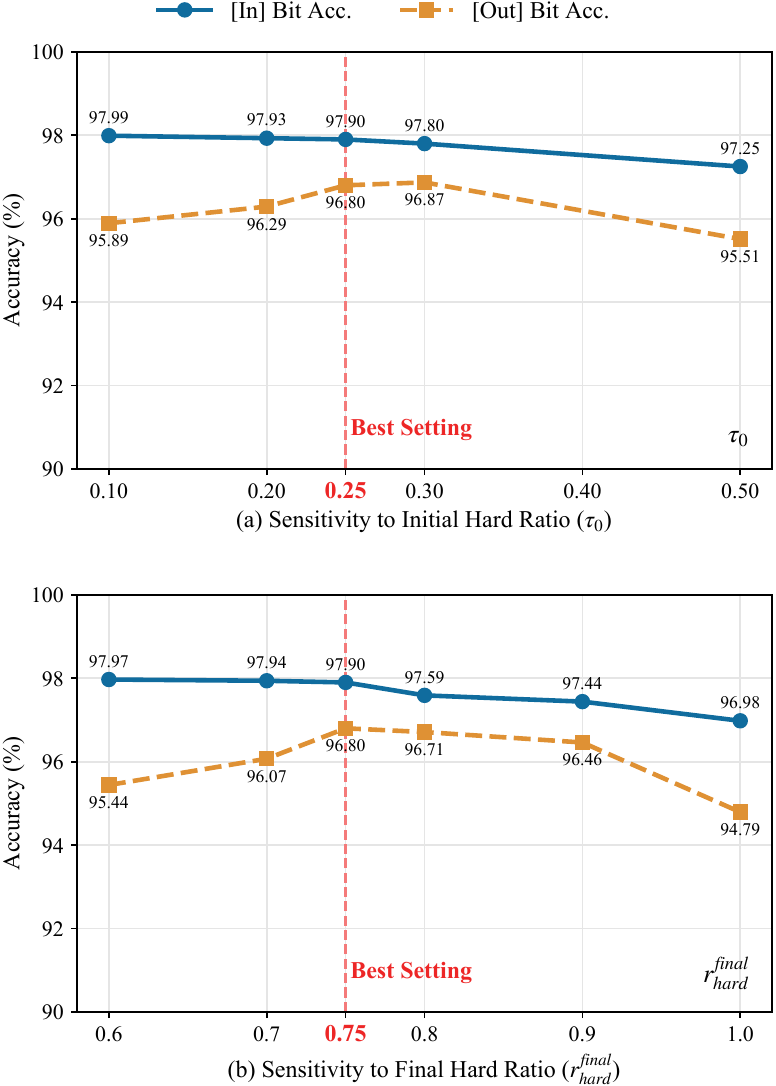}
\caption{\label{fig_hms_sensitivity} Sensitivity to hard-message sampling ratios $\tau_0$ and $r_{\text{hard}}^{\text{final}}$ under the 64-bit capacity setting.} 
\end{figure}

\subsection{Sensitivity Analysis}

\noindent \textit{1) Sensitivity to the bit compression rate ($n$).} Table~\ref{tab:sensitivity1} shows how the bit compression rate $n$ affects decoding accuracy across payload sizes, revealing a trade-off between token-budget relaxation and chunk prediction difficulty. For 16 bits, $n=1$ performs best (99.85\%), indicating compression is unnecessary when the payload already fits the token budget. For 32–96 bits, $n=2$ consistently achieves the highest accuracy (99.67\%, 98.87\%, 97.35\%, and 93.97\%). At 128 bits, stronger compression becomes beneficial, with $n=4$ reaching 90.59\%. Larger compression ($n=8$) consistently underperforms due to the expanded chunk space ($2^8$), which increases prediction difficulty. Overall, moderate compression is preferable in most cases, while stronger compression helps only at very high payloads.

\begin{table}[!ht]
    \centering
    \caption{Sensitivity to the Bit Compression Rate.}
    \label{tab:sensitivity1}
    \begin{tabular}{l |cccc}
            \toprule
            \multirow{2}{*}{Bit} & $n=1$  & $n=2$ & $n=4$ & $n=8$ \\
            \cmidrule(lr){2-5}
            & Bit-Acc & Bit-Acc &  Bit-Acc & Bit-Acc \\
            \midrule
            16 & \cellcolor{black!10}\textbf{99.85} & 99.10  & 97.97 & 96.42 \\
            32 & 99.49 & \cellcolor{black!10}\textbf{99.67}  & 98.06 & 96.90 \\
            48 & 98.85 & \cellcolor{black!10}\textbf{98.87}  & 97.91 & 95.39  \\
            64 &  96.71 & \cellcolor{black!10}\textbf{97.35}  & 95.01 & 93.88 \\
            96 & \ding{53} & \cellcolor{black!10}\textbf{93.97} & 93.19 & 91.65 \\
            128 & \ding{53} & 89.93 & \cellcolor{black!10}\textbf{90.59} & 89.40 \\
            \bottomrule
        \end{tabular}
\end{table}

\noindent \textit{2) Sensitivity to the number of decoder groups  ($G$).} Table~\ref{tab:sensitivity2} examines the effect of the group number $G$ in the bit branch and shows a milder impact than the compression rate $n$. For 16 bits, a single group performs best, indicating that low-capacity messages do not benefit from additional grouping. For larger payloads (32–128 bits), $G=4$ consistently achieves the highest accuracy (99.67\%, 98.87\%, 97.35\%, 93.97\%, and 90.59\%). Increasing $G$ to 8 or 16 brings no further gains, suggesting that overly fine-grained grouping fragments bit-level dependencies. Therefore, we use $G=4$ by default for high-capacity settings and $G=1$ for 16-bit messages. \par\vspace{0.8\baselineskip}\noindent

\begin{table}[!ht]
    \centering
    \caption{Sensitivity to the Number of Decoder Groups.}
    \label{tab:sensitivity2}
    \begin{tabular}{l |cccc}
            \toprule
            \multirow{2}{*}{Bit} & $G=1$  & $G=4$ & $G=8$ & $G=16$ \\
            \cmidrule(lr){2-5}
            & Bit-Acc & Bit-Acc &  Bit-Acc & Bit-Acc \\
            \midrule
            16 & \cellcolor{black!10}\textbf{99.85} & 99.65 & 99.07 & 98.38 \\
            32 & 99.63 & \cellcolor{black!10}\textbf{99.67} & 99.60 & 99.45\\
            48 & 98.69 & \cellcolor{black!10}\textbf{98.87} & 98.51 & 98.08 \\
            64 & 97.02 & \cellcolor{black!10} \textbf{97.35} & 96.88 & 96.28 \\
            96 & 92.85 & \cellcolor{black!10} \textbf{93.97} & 93.15 & 93.01 \\
            128 & 87.57 & \cellcolor{black!10} \textbf{90.59} & 90.03 & 89.36 \\
            \bottomrule
        \end{tabular}
\end{table}

\noindent \textit{3) Sensitivity to hard-message sampling ratios.}
We control the hard-message ratio using a linear warm-up schedule,
$r_{\mathrm{hard}}=\min(1.0,\tau_0+\alpha \cdot e)$.
For clarity, the pre-freeze schedule is parameterized by the initial hard ratio $\tau_0$ and its nominal endpoint $r_{\mathrm{hard}}^{\mathrm{final}}$, with $\alpha=(r_{\mathrm{hard}}^{\mathrm{final}}-\tau_0)/E_{\mathrm{freeze}}$.
These parameters govern the transition from broad message exploration to hard-message concentration during decoder pre-training. Under the 64-bit setting, we sweep $\tau_0 \in \{0.1,0.2,0.25,0.3,0.5\}$ and $r_{\mathrm{hard}}^{\mathrm{final}} \in \{0.6,0.7,0.75,0.8,0.9,1.0\}$. As shown in Fig.~\ref{fig_hms_sensitivity}, performance remains stable across a wide range of settings, with only minor variation in both \textit{[In]} and \textit{[Out]} accuracy. Results degrade when the schedule becomes overly aggressive, for example, when starting with too many hard samples or pushing the final ratio close to 1.0, which reduces message diversity and harms unseen-message recovery. The default setting $(\tau_0=0.25,\, r_{\mathrm{hard}}^{\mathrm{final}}=0.75)$ provides the best overall trade-off.

\section{Conclusion}
\label{sec:conclusion}
We presented BitC-3DGS, a high-capacity watermarking framework for 3D Gaussian Splatting that overcomes the 77-bit limit of CLIP-based bit-to-token encoding via bit-compressed tokenization. To ensure reliable decoding, we introduced a dual-branch decoder that jointly performs chunk-level and bit-level prediction, together with a hard-message sampling strategy that improves training coverage and reduces bias toward seen messages. Experiments on Blender and LLFF show that BitC-3DGS achieves strong recovery accuracy while preserving high rendering fidelity. It scales effectively to 96-bit and 128-bit payloads and reduces the seen–unseen performance gap compared with prior methods. Overall, BitC-3DGS demonstrates that high-capacity semantic watermarking for 3DGS can be achieved without compromising visual quality.

\bibliographystyle{IEEEtran}
\bibliography{Manuscript}{}

\end{document}